\documentclass{article}

\PassOptionsToPackage{numbers, compress}{natbib}

 \usepackage[preprint]{neurips_2024}

\usepackage{comment}
\usepackage[utf8]{inputenc} 
\usepackage[T1]{fontenc}    
\usepackage{hyperref}       
\usepackage{url}            
\usepackage{booktabs}       
\usepackage{amsfonts}       
\usepackage{nicefrac}       
\usepackage{microtype}      
\usepackage{xcolor}
\usepackage{graphicx}
\usepackage{amsmath}
\usepackage{amsfonts}
\usepackage{enumitem}
\usepackage{subcaption}
\usepackage{natbib}
\usepackage{float}
\usepackage[font=scriptsize, labelfont=bf]{caption}
\graphicspath{{figures/}}

\title{Leveraging Open-Source Large Language Models for encoding Social Determinants of Health using an Intelligent Router}

\author{%
Akul Goel$^{1}$ \quad Surya Narayanan Hari$^{1}$ \quad Belinda Waltman$^{2}$ \quad Matt Thomson$^{1}$ \\
\quad $^{1}$ California Institute of Technology \\
\quad $^{2}$ Los Angeles County Department of Health Services \\
\texttt{\{agoel,shari,mthomson\}@caltech.edu}}

\begin{document}

\maketitle

\begin{abstract}

  Social Determinants of Health (SDOH), also known as Health-Related Social Needs (HRSN), play a significant role in patient health outcomes. The Centers for Disease Control and Prevention (CDC) introduced a subset of ICD-10 codes called Z-codes to recognize and measure SDOH. However, Z-codes are infrequently coded in a patient’s Electronic Health Record (EHR), and instead, in many cases, need to be inferred from clinical notes. Previous research has shown that large language models (LLMs) show promise on extracting unstructured data from EHRs, but it can be difficult to identify a single model that performs best on varied coding tasks. Further, clinical notes contain protected health information posing a challenge for the use of closed-source language models from commercial vendors. The identification of open source LLMs that can be run within health organizations and exhibit high performance on SDOH tasks is an important issue to solve. Here, we introduce an intelligent routing system for SDOH coding that uses a language model router to direct medical record data to open-source LLMs that demonstrate optimal performance on specific SDOH codes. The intelligent routing system exhibits state of the art performance of 96.4\% accuracy averaged across 13 codes, including homelessness and food insecurity, outperforming with closed models such as GPT-4o. We leveraged a publicly-available, deidentified dataset of medical record notes to run the router, but we also introduce a synthetic data generation and validation paradigm to increase the scale of training data without needing privacy-protected medical records. Together, we demonstrate an architecture for intelligent routing of inputs to task-optimal language models to achieve high performance across a set of medical coding sub-tasks.
  
\end{abstract}

\section{Introduction}

Social determinants of health (SDOH) are defined as non-medical factors that strongly influence health outcomes. These are also known as Health-Related Social Needs (HRSN). SDOH are increasingly regarded as significant risk factors across a broad range of health outcomes and generate measures of well-being \cite{solar2010conceptual}. SDOH generally include factors like economic stability and access to resources such as food, housing, education, health care.  Ideally a holistic, whole-person approach to physical and behavioral health should include a focus on these social determinants of health because acute or chronic conditions cannot be treated in a vacuum. For example, it may be challenging for a low-income cancer patient to focus on treatment when unpaid sick days could exacerbate their food and housing insecurity \cite{waltman2024margins}. Given the central importance of SDOH, there is an increased focus on screening for SDOH and applying interventions in the form of social services to impact SDOH. The Centers for Medicare and Medicaid Services (CMS) introduced a subset of ICD-10 codes called Z-codes in an attempt to officially recognize and measure SDOH in the health care system \cite{cms2023zcode}. However, an important limitation for SDOH scoring is related to data - historically, many of the notes related to SDOH domains were free-texted into the Social History section of the EMR, so it is often difficult to extract for both reporting and clinical intervention purposes. 

Large Language Models (LLMs) show promise for extracting data from unstructured medical notes. They have already been used on a variety of clinical tasks such as predicting readmission rates from Electronic Health Records (EHRs), supporting clinical decision-making\cite{clusmann_future_2023}\cite{omiye_large_2024}\cite{mehandru_evaluating_2024},and identifying generally ’adverse’ SDOH codes from EHR notes \cite{guevara_large_2024}. However, achieving high accuracy results generally use closed-model architectures like GPT4 or Claude, or require significant fine-tuning \cite{guevara_large_2024}\cite{lybarger_annotating_2021}\cite{omiye_large_2024}. Closed-model architectures are problematic for the SDOH use cases because they are LLMs whose training data, weights, and prompting strategy are proprietary, they are hosted on industrial cloud computing infrastructure, and they require the transmission of protected health information (PHI). As an alternative to closed source models, open-source language models have proliferated in recent years with models like LLAMA and Mistral achieving performance on par with closed source models \cite{touvron_llama_2023, jiang_mistral_2023}. However, there are currently over 100,000 open-source language models on the Hugging Face platform, and it is difficult to discover open source LLMs that might be optimal for a given task \cite{hari_herd_2023, hari_tryage_2023}. 
Further, fine-tuning open-source models requires valuable clinical data where a paucity of data related to limitations around data privacy and also data quality issues can cause model performance to suffer \cite{cook_quality_2021}. 

Here we introduce an intelligent routing architecture that leverages multiple open-source LLMs that do not need to be fine-tuned to achieve state of the art accuracy in the SDOH coding task. Our router not only navigates the problem of open-source models being subject to a variety of training architectures and possessing mixed quality data \cite{wang_openchat_2024}\cite{bommasani_opportunities_2021}\cite{jin_dataless_2022},but it also leverages these differences by selecting LLMs whose training data align more effectively with specific SDOH codes. We also introduce a new synthetic data generation and validation paradigm to address the issue of lack of high-quality clinical data that is often hidden behind privacy protected medical records. Our router system trained on this synthetic data, without fine-tuning or using closed-source models, is able to achieve 96.4\% accuracy, on par with much bigger, top-of-the-line models like GPT-4o.

\section{Methods}
\begin{figure*}[t!]
    \centering
    \begin{subfigure}[t]{0.5\textwidth}
        \centering
        \includegraphics[width=\textwidth]{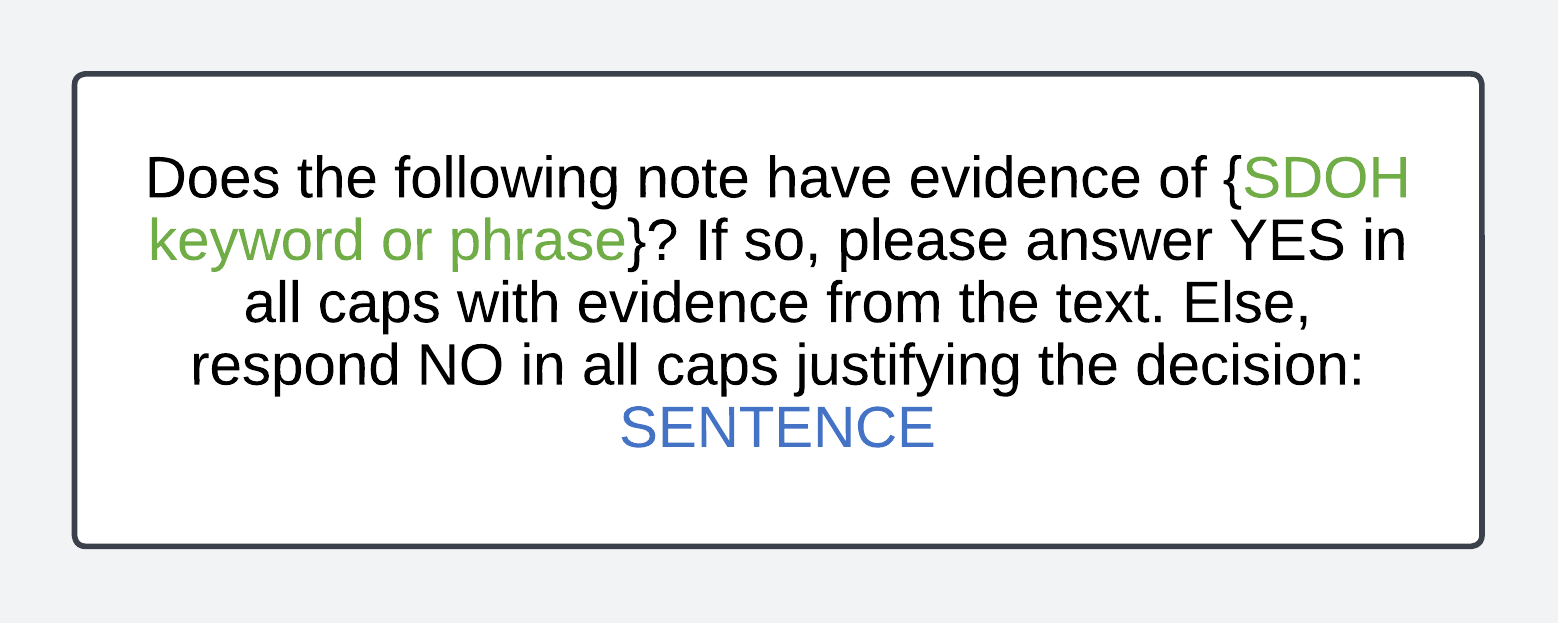}
        \caption{}
    \end{subfigure}%
    ~ 
    \begin{subfigure}[t]{0.5\textwidth}
        \centering
        \includegraphics[width=\textwidth]{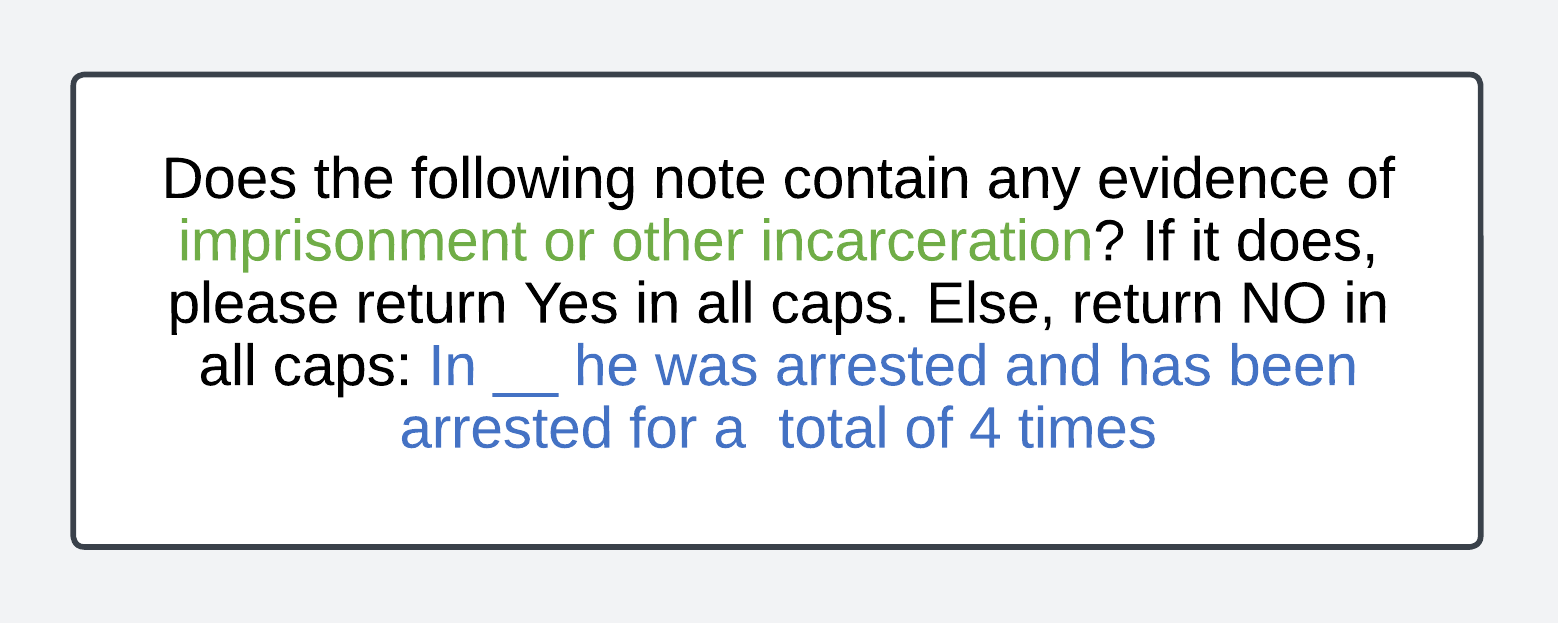}
        \caption{}
    \end{subfigure}
    \caption{\textbf{Models were prompted with a sentence from the medical note to classify.} a) The above figure shows the prompt template used to prompt all models on all SDOH codes. The specific SDOH we were aiming to classify was input into the {SDOH keyword or phrase} (green) field, and a sentence from the medical note the model was to classify was placed in the SENTENCE (blue) field. b) This figure shows a specific example of a prompt for SDOH code 'imprisonment or other incarceration' (green), with a sentence (blue) from the medical note corresponding to it.}
    \label{figure:prompt}
\end{figure*}

Our data came from patient electronic health records (EHRs) stored in the MIMIC-III dataset, which comprises over 40,000 publicly-available, de-identified patient records and more than two million clinical notes collected between 2001 and 2012 \cite{johnson_mimic-iii_2015}. We created an enriched data set with notes that contained a “Social History” section, which distilled the dataset down to approximately 8000 notes. We then randomly selected 500 of those notes. For a proof of concept, we initially tested 5 factors including homelessness, food insecurity, imprisonment or other incarceration, low income, and marital estrangement, relatives needing care, and unemployment. A professional medical coder applied Z-codes to 500 medical notes by performing a manual keyword analysis to each note, indicating information verbatim from the medical notes that corresponded to a specific SDOH code. Noticing that these codes were assigned based on a phrase from the note, we determined that a sentence had enough information for an LLM to appropriately identify the codes bearing evidence in the medical records. We prompted a number of open-source models with the Together API using the prompt template in Figure \ref{figure:prompt}.

\begin{figure}[H]
    \centering
    \includegraphics[width = \linewidth]{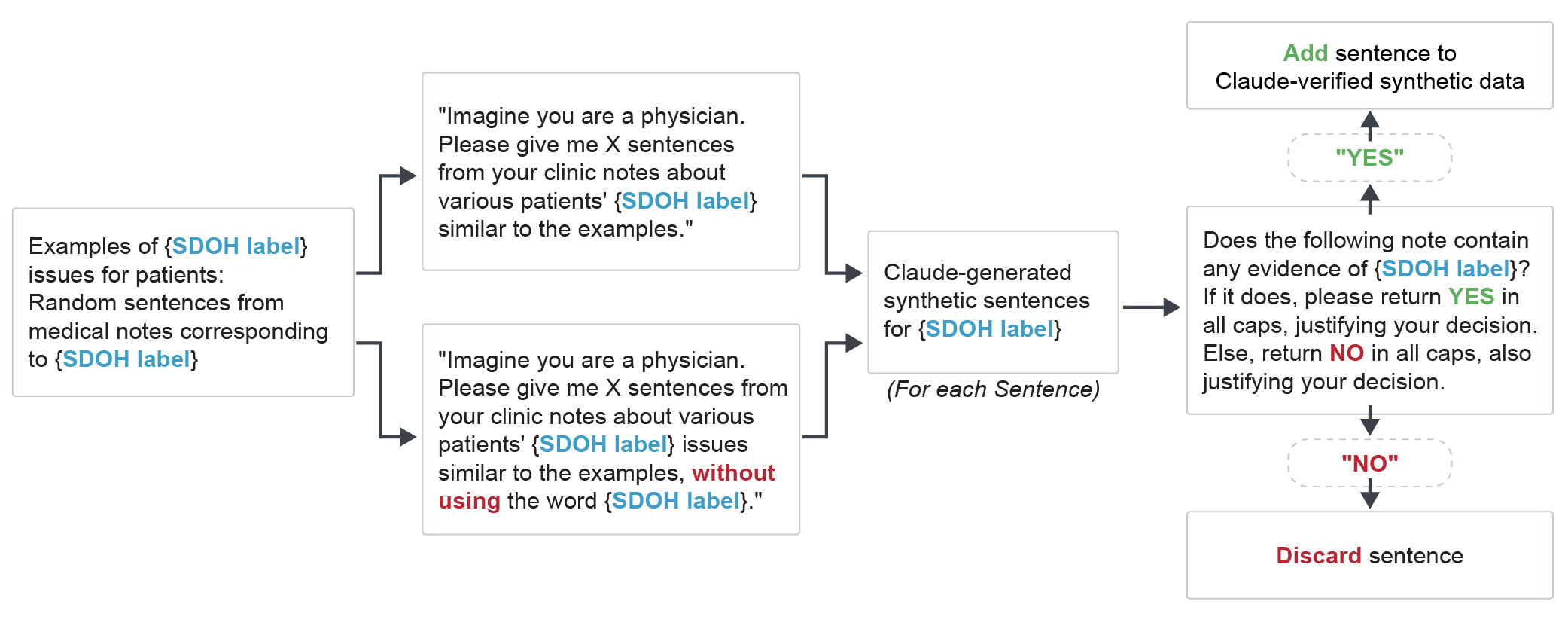}
    \caption{\textbf{Claude-3-opus generated and verified synthetic data.} Figure shows flow chart of synthetic data generation and verification by Claude-3-opus. Random sentences corresponding to a specific SDOH code were given as examples from labeled medical notes. Based on these examples, Claude was asked to generate synthetic sentences via two different prompts, one explicitly asking the model to not use the SDOH keyword for complexity. Each synthetic sentence was subsequently checked by Claude for evidence of the specific SDOH code and sentences that did not pass were discarded. Sentences that passed the verification were added to an SDOH’s synthetic data collection. Random sentences for examples were picked from 500 labeled medical notes from the MIMIC-III dataset \cite{johnson_mimic-iii_2015}. }
    \label{fig:synthetic_data}
\end{figure}

Even after manually-coding 500 medical notes, sentences with evidence of specific SDOH codes were sparse (<1\% of the total number of sentences in the 500 notes). In order to address this, we introduced a novel synthetic data generation and validation paradigm, outlined in \ref{fig:synthetic_data}. Our synthetic data generation scheme involved passing in random sentences corresponding to a specific SDOH code from the medical note into Claude-3-opus as references. We were able to use closed-source models like Claude-3-opus in this generation scheme because the dataset was deidentified and did not include protected health information (PHI). Using two distinct prompts, one explicitly asking Claude not to use the SDOH keyword or phrase for complexity, we generated a couple hundred synthetic sentences corresponding to SDOH codes. We subsequently had Claude verify the synthetic sentences it generated by asking Claude to determine if the generated synthetic sentence had evidence of a specific SDOH code, dropping the sentences that did not pass this test. As evidenced in Figure \ref{figure:dropped_data}, Claude showed an ability to self-edit its own generated data. This scheme ultimately resulting in a validation set distribution seen in Figure \ref{figure:data_distributions}. 

\begin{figure}[H]
    \centering
    \begin{tabular}{|c|c|c|c|}
    \hline
    \textbf{SDOH Code} & \textbf{Gold Data} & \textbf{Synthetic Data} & \textbf{Negative Data} \\ \hline
    Homelessness & 28 & 318 & 704 \\ \hline
    Food Insecurity & 2 & 349 & 702 \\ \hline
    Imprisonment or Other Incarceration & 17 & 313 & 660 \\ \hline
    Marital Estrangement & 53 & 312 & 730 \\ \hline
    Relative Needing Care & 3 & 320 & 646 \\ \hline
    \end{tabular}
    \caption{\textbf{Test Set displayed approximately 33\%/67\% splits for each code.} Table shows data distribution across 5 SDOH codes, with various numbers of gold data (sentences from medical note containing SDOH code), synthetic data (sentences generated and validated by Claude in Fig. \ref{fig:synthetic_data} containing SDOH code), and negative data (random subset of sentences from medical note not containing SDOH code). This validation set ultimately contained approximately 33\% positive label (defined as gold + synthetic data) and 67\% negative label (negative data) splits for each SDOH task, with around 1000 total sentences for each code. Gold data and negative data were sentences pulled from 500 labeled medical notes from the MIMIC-III dataset \cite{johnson_mimic-iii_2015}. }
    \label{figure:data_distributions}
\end{figure}

The validation set consisted of approximately 33/67\% splits of positive/negative labels, with approximately 1000 sentences for each SDOH code. Positive labels consisted of both data containing evidence of an SDOH code from the medical note (gold data) and verified synthetic data (generated and verified by scheme shown in Figure \ref{fig:synthetic_data}). Negative data consisted of sentences from medical notes not containing evidence of an SDOH code. This was the dataset used both to train the router, and was tested against to ultimately determine the performance seen in Figure \ref{figure:comparison_plot}. Interestingly, we observed that Claude had the ability to verify synthetic data it generated itself, resulting in ‘dropped’ data numbers seen in Figure \ref{figure:dropped_data}. 

\begin{figure}[H]
    \centering
    \begin{tabular}{|c|c|}
    \hline
    \textbf{SDOH Code} & \textbf{Claude Dropped} \\ \hline
    Homelessness & 12 \\ \hline
    Food Insecurity & 51 \\ \hline
    Imprisonment or Other Incarceration & 7 \\ \hline
    Marital Estrangement & 8 \\ \hline
    Relative Needing Care & 0 \\ \hline
    \end{tabular}
    \caption{\textbf{Verification scheme resulted in Claude dropping synthetic data.} Table for 5 SDOH codes showing numbers of synthetic data originally dropped by Claude in scheme shown in Fig. \ref{fig:synthetic_data}. Data distributions after “Claude Dropped” is shown in Fig. \ref{figure:data_distributions} and was the dataset used to generate Figs. \ref{figure:comparison_plot}}
    \label{figure:dropped_data}
\end{figure}

\begin{figure}[H]
    \centering
    \includegraphics[width = \linewidth]{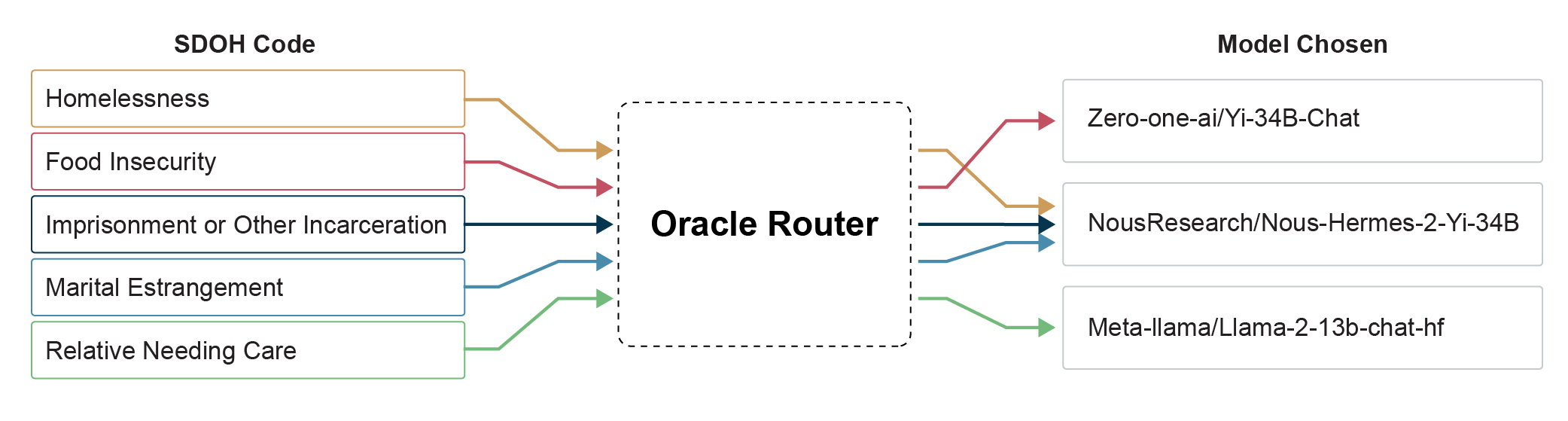}
    \caption{\textbf{Router identifies best model for given SDOH code.} Router model architecture. Oracle router takes in keyword or phrase from given SDOH and outputs best model for the task. Figure shows color coded arrow from SDOH code to the model chosen by the router (ie, for food insecurity, Zero-one-ai/Yi-34B-Chat is considered the best model).}
    \label{fig:router_schematic}
\end{figure}

We used the synthetic data generated by the paradigm introduced in Figure \ref{fig:synthetic_data} in order to train an “Oracle Router”. This intelligent router (whose schematic is shown in Figure \ref{fig:router_schematic}) takes in a SDOH keyword or phrase, and outputs the open-source LLM that will perform the best on a given task. The router was trained on both synthetic data (scheme shown in Figure \ref{fig:synthetic_data}) and labelled sentences from the MIMIC-III dataset \cite{johnson_mimic-iii_2015}, and executes its function by choosing the model that maximizes accuracy for a given SDOH coding task.  

After evaluating the model’s performance on the preliminary dataset of 1,000 sentences, we sought to expand the range of SDOH codes considered. We implemented an inter-model annotation scheme (in-stead of human medical coders, due to the high initial accuracy in preliminary results) to annotate the remaining 8,000 notes for 13 SDOH factors: Illiteracy or low-level literacy, Unemployment, Home-lessness, Food insecurity, Low income, Insufficient social insurance or welfare support, Problems related to living alone, Abuse in childhood, Problems in relationships with a spouse or partner, Disap-pearance or death of a family member, Marital estrangement, Relatives needing care, Alcoholism or drug addiction in the family, and Imprisonment or other incarceration.

In total, our inter-model annotation scheme worked as follows. We used two closed-source models (GPT-4o and Claude Opus 3) and Nous-Hermes-2-Yi-34B as an open-source model to resolve any disagreements. If two out of the three models agreed on a sentence’s label, we assigned that label accordingly. Nous-Hermes-2-Yi-34B was selected as the open-source model due to its strong performance in the preliminary study. In practice, this was implemented as follows: all 8,000 notes were first processed by Nous-Hermes, with each note being evaluated for the 13 SDOH codes. Sentences identified as positive examples (i.e., containing evidence of a specific code) by Nous-Hermes were then reviewed by GPT-4o. If GPT-4o agreed with the annotation, the sentence was labeled as containing evidence for that specific code. However, if GPT-4o disagreed, the sentence was further evaluated by Claude Opus 3, which acted as the tiebreaker.

\begin{figure}[H]
    \centering
    \includegraphics[width = \linewidth]{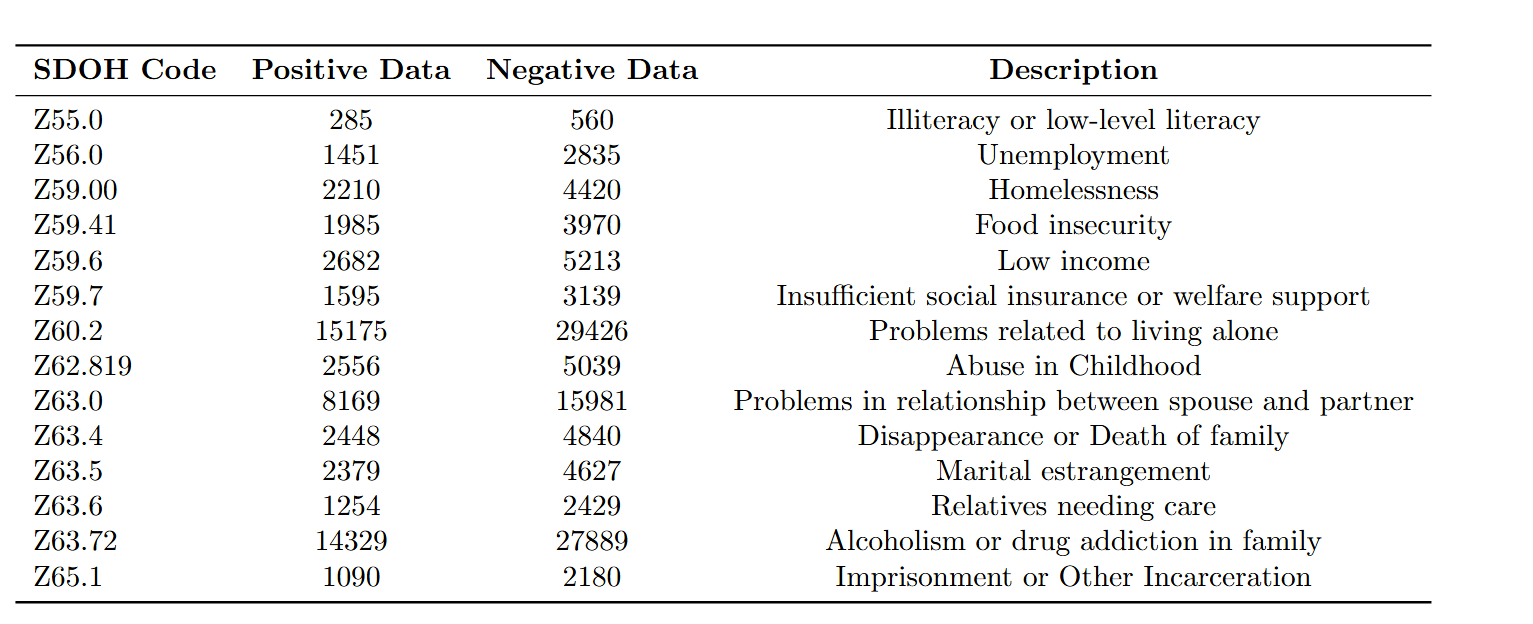}
    \caption{\textbf{Test Set Distribution After Inter-Model Annotation Scheme}}
    \label{fig:total_table}
\end{figure}

Labeling negative sentences followed a similar process. In totality, this netted a test set for our routing system as can be seen in Figure \ref{fig:total_table}.

The metrics we used were the following: We evaluated models both based on accuracy and F1 score. In order to calculate these values, we define true positive (TP), true negative (TN), false positive (FP), and false negative (FN) as follows:

\begin{itemize}
    \item \textbf{TP:} Sentence or note correctly identified as containing evidence of SDOH code
    \item \textbf{TN:} Sentence or note correctly identified as not containing evidence of SDOH code
    \item \textbf{FP:} Sentence or note incorrectly identified as containing evidence of SDOH code
    \item \textbf{FN:} Sentence or note incorrectly identified as not containing evidence of SDOH code
\end{itemize}

Thus, the formulas for accuracy and F1 score are as follows:

\begin{align}
\text{Accuracy} &= \frac{\text{TP} + \text{TN}}{\text{TP} + \text{TN} + \text{FP} + \text{FN}} = \frac{\text{TP} + \text{TN}}{\text{Total}} \\
\text{Precision} &= \frac{\text{TP}}{\text{TP} + \text{FP}} \\
\text{Recall} &= \frac{\text{TP}}{\text{TP} + \text{FN}} \\
\text{F1} &= \frac{2 \times \text{Precision} \times \text{Recall}}{\text{Precision} + \text{Recall}}
\end{align}

\section{Results and Discussion}
\subsection*{Identification of open-source models for specific SDOH codes}

Language models are trained on distinct data domains and show differential performance on tasks including the identification of specific SDOH codes \cite{hari_herd_2023, hari_tryage_2023}. SDOH codes span a variety of underlying topics from housing to family relationships to incarceration status. Hence, coding is well-suited to an ensemble approach where expert models are used to analyze clinical notes for specific SDOH factors. Therefore, we developed an intelligent routing system for SDOH coding that consists of a router model and an ensemble of down-stream open-source models that show expert performance on coding for specific SDOH factors.

\begin{figure}[H]
    \centering
    \includegraphics[width=\linewidth]{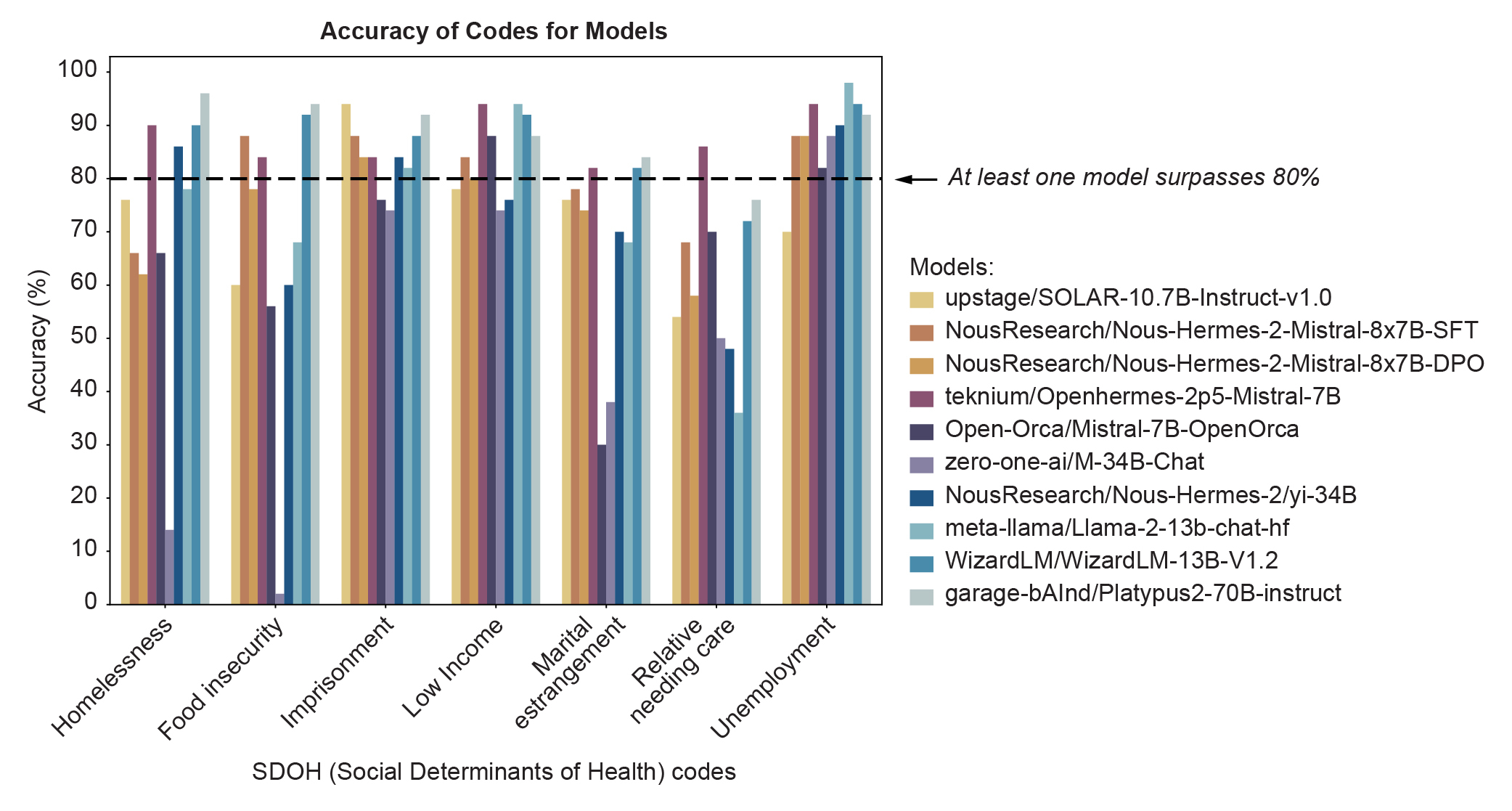}
    \caption{\textbf{LLMs exhibit differential performance on SDOH coding task}. Plot shows performance in accuracy of ten different open-source LLMs on seven SDOH codes, where accuracy is defined as (True Positives + True Negatives) / (Total Number of Notes). True positive is defined as correctly identifying a note as needing a specific SDOH code, and true negative is defined as correctly identifying a note as not needing a specific SDOH code. Models are indicated by legend (in distinct colors). Different models emerge as the best model depending on the specific SDOH task, with at least one model achieving 80\% accuracy on each code. Analysis was performed on 50 medical notes from the MIMIC-III dataset \cite{johnson_mimic-iii_2015}.}
    \label{fig:large_plot}
\end{figure}

To identify optimal open-source models, for our initial proof-of-concept we generated a dataset that contained both 500 medical notes from the MIMIC-III dataset \cite{johnson_mimic-iii_2015}, as well as synthetic examples generated using an LLM. We, first, analyzed the performance of a set of open-source language models on each of seven codes, as well as synthetic examples generated using an LLM. We, first, analyzed the performance of a set of open source language models on each of seven codes (Figure \ref{fig:large_plot}), and, then, trained a router to route incoming coding tasks to the optimal down-stream model (Figure \ref{fig:router_schematic}).

Figure \ref{fig:large_plot} shows the performance of ten open-source large language models (LLMs) across seven social determinants of health code (SDOH) on a small dataset of 50 notes from the MIMIC-III database \cite{johnson_mimic-iii_2015}. From this graph, we were able to identify at least one model that achieved >80\% accuracy on each code, indicating overall that open-source LLMs have the ability to extract SDOH information from unstructured medical charts. However, we observed variable performance on these notes, where a different model performs the best on a given code. We also see that some codes, such as unemployment, demonstrate high accuracy across models, while others, like relative needing care, possess more variable and lower overall accuracy scores. 

\begin{figure}[h]
    \centering
    \includegraphics[width= \linewidth]{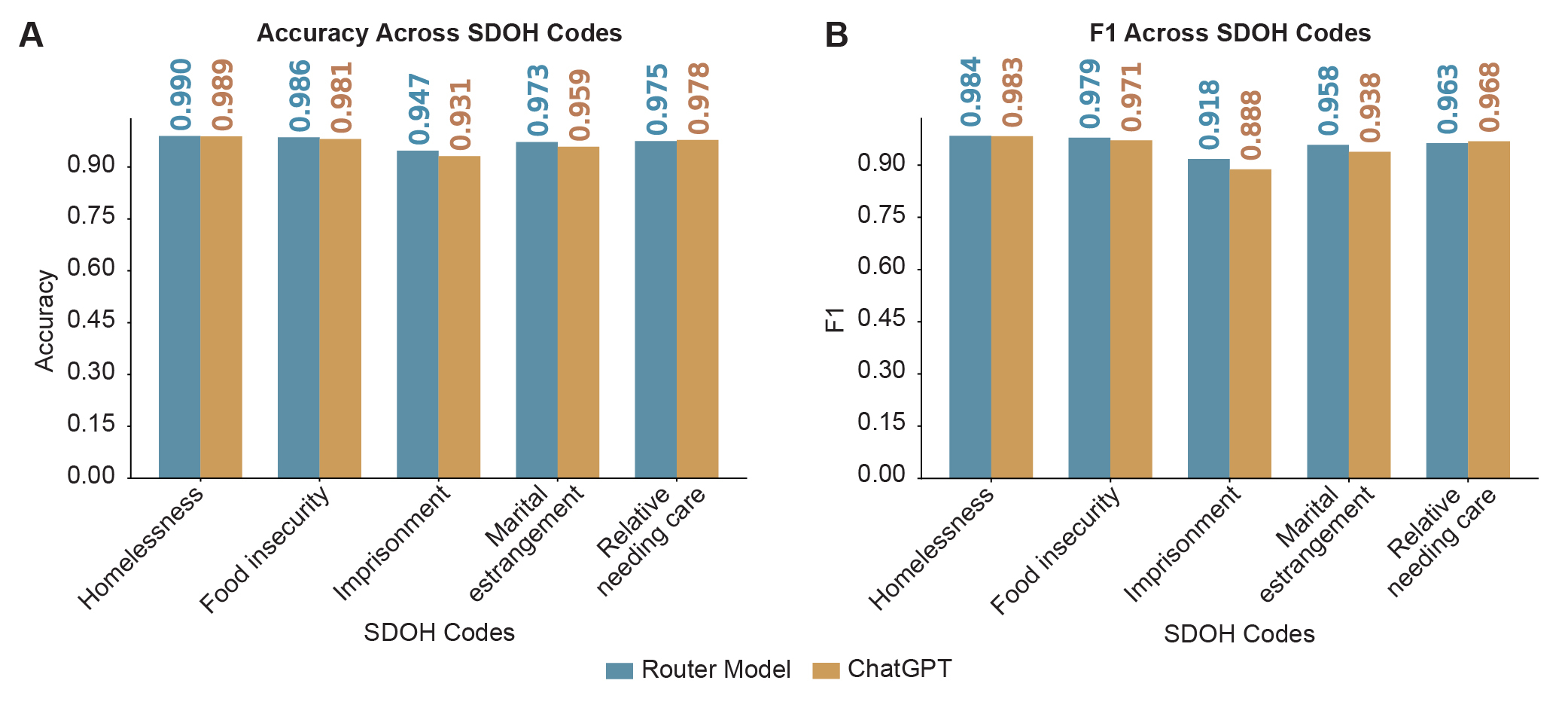}
    \caption{\textbf{Router chosen model exhibits high accuracy across codes, with comparable performance to GPT-4o.}  a) Performance in accuracy of router chosen model (blue) and GPT-4o (gold) against 5 SDOH codes on the dataset described in Fig. \ref{figure:data_distributions}. Accuracy is defined as (True Positive + True Negative) / (Total Number of Sentences) where true positives are sentences correctly identified as having evidence of specific SDOH data, and true negatives are defined as sentences correctly identified as having no evidence of specific  SDOH data. Router exhibits comparable accuracy to GPT-4o, slightly beating it in 4 of the 5 codes.  b) Performance in F1 score of router chosen model (blue) and GPT-4o (gold) against 5 SDOH codes on the dataset described in Fig. \ref{figure:data_distributions}. The F1 score is defined as (2*Precision*Recall) / (Recall + Precision), where precision is calculated by (True Positive)/ (True Positive + False Positive), and recall is calculated as (True Positive)/ (True Positive + False Negative). True positive is defined as correctly identifying a sentence as needing a specific SDOH code, and true negative is defined as correctly identifying a sentence as not needing a specific SDOH code. False positive is defined as incorrectly identifying a sentence as needing a specific SDOH code, and false negative is defined as incorrectly identifying a sentence as not needing a specific SDOH code. Router chosen model exhibits F1 scores to GPT-4o, slightly beating it in 4 of the 5 codes. Note that the label “imprisonment” was shortened from “imprisonment or other incarceration” on both graphs for visualization purposes. }
    \label{figure:comparison_plot}
\end{figure}

Only a few sentences from the medical notes (<1\% of the sentences) contained  evidence of specific SDOH data, so we generated and validated synthetic data. Furthermore, after manually labeling 500 medical notes (by a professional medical coder) from the MIMIC-III dataset\cite{johnson_mimic-iii_2015}, we distilled our validation set to test only 5 SDOH codes (homelessness, food insecurity, imprisonment or other incarceration, marital estrangement, and relative needing care) for a proof of concept. Training a router to pick a best model for a given SDOH factor from the medical note, we observed the performance shown in Figure \ref{figure:comparison_plot}. On a 1000 sentence validation set, containing sentences from 500 labeled medical notes found in the MIMIC-III dataset \cite{johnson_mimic-iii_2015} and synthetic data, our router picked a model that achieved >90\% accuracy and F1 score for every single code. The highest performing code was homelessness, where the router picked the model identified this code with 99.0\% accuracy (0.984 F1 score). The lowest performing  code was imprisonment or other incarceration, at 94.7\% accuracy (0.918 F1 score). Comparing the performance of the router picked model to GPT-4o, our chosen model beats that of GPT for 4 of the 5 codes, only exhibiting marginally worse performance for the relative needing care SDOH factor (97.5\% vs 97.8\% accuracy for our router picked model vs GPT-4o).

\begin{figure}[H]
    \centering
    \begin{tabular}{|c|c|}
    \hline
    \textbf{SDOH Code} & \textbf{Model Chosen} \\ \hline
    Homelessness & NousResearch/Nous-Hermes-2-Yi-34B \\ \hline
    Food Insecurity & Zero-one-ai/Yi-34B-Chat \\ \hline
    Imprisonment or Other Incarceration & NousResearch/Nous-Hermes-2-Yi-34B \\ \hline
    Marital Estrangement & NousResearch/Nous-Hermes-2-Yi-34B \\ \hline
    Relative Needing Care & Meta-llama/Llama-2-13b-chat-hf \\ \hline
    \end{tabular}
    \caption{\textbf{Router chooses different model based on SDOH code.} Table showing model chosen by router for each of 5 SDOH codes. Their performances on the test set are represented in the ‘router model’ bar graph in Fig. \ref{figure:comparison_plot}}
    \label{figure:model_chosen}
\end{figure}

In Figure \ref{figure:model_chosen}, we see that the router selected a different best open-source performing model based on the SDOH code to achieve the performance seen in Figure \ref{figure:comparison_plot}. The most common model picked was NousResearch/Nous-Hermes-2-Yi-34B, chosen for 3 of the 5 SDOH codes. The LLMs Zero-one-ai/Yi-34B-Chat and Meta-llama/Llama-2-13b-chat-hf were chosen for one code each. All of these models were evaluated without fine-tuning, are completely open-source, and can be found on HuggingFace.

\begin{figure}[h]
\centering
\includegraphics[width= \linewidth]{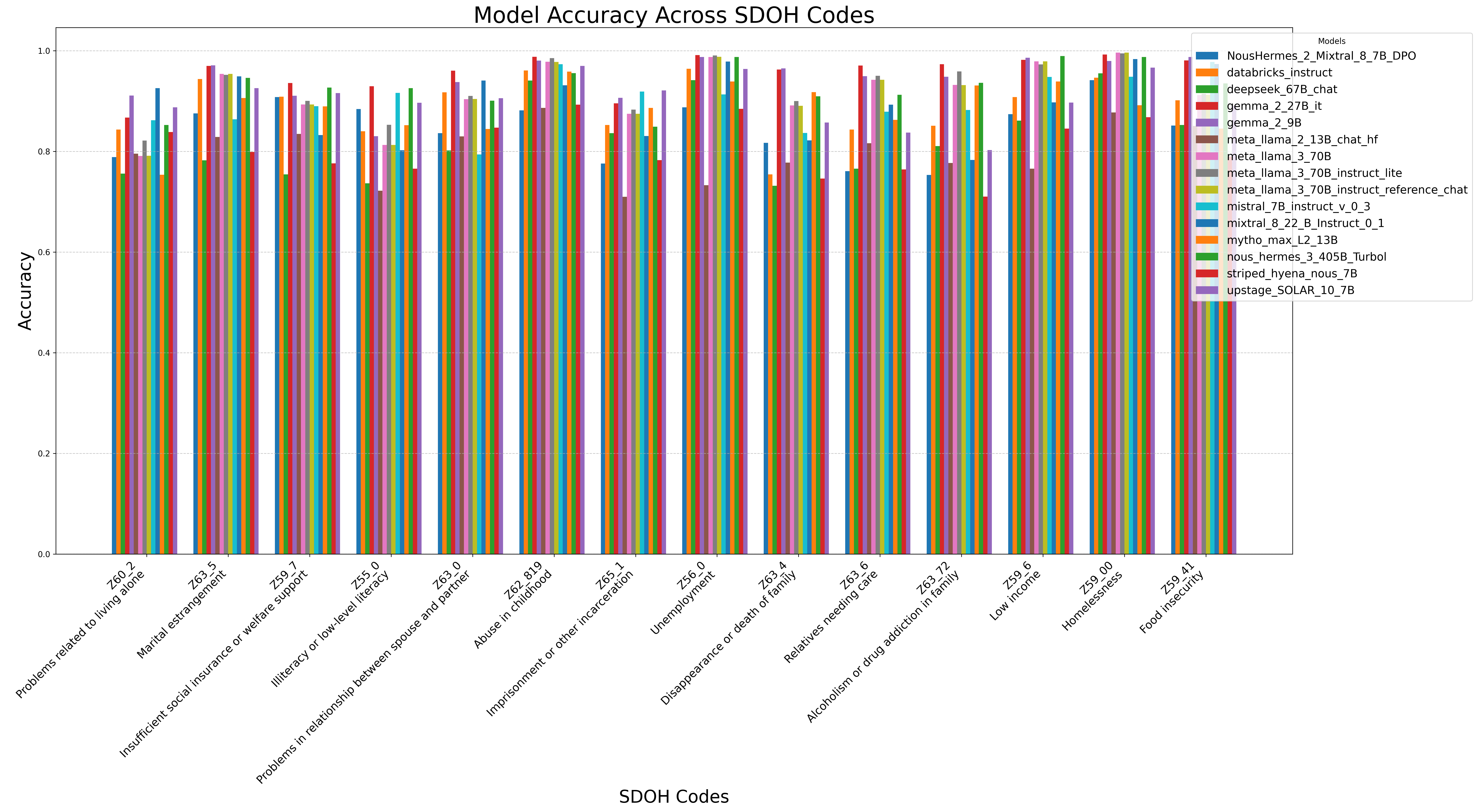}
\caption{\textbf{Open-source Models Show Differential Performance Across Medical Notes}. Expanding our dataset to 8000 medical notes and 13 SDOH codes, we see differential performance across 15 open-source models.}
\label{figure:total_plot}
\end{figure}

Using the results from the 500 notes as a proof of concept, we expanded our final analysis to over 8000 notes from the MIMIC-III dataset \cite{johnson_mimic-iii_2015} and evaluated around 15 open-source models across 13 SDOH codes (outlined in the Methods). The expanded dataset allowed us to assess model performance on a more diverse and representative set of clinical notes, revealing substantial variability in accuracy across different models and codes. Also, with the increase in number of notes, for this final comprehensive dataset, we no longer used synthetic data to test the router (only using synthetic data to train the routing system). Furthermore, the proof-of-concept study revealed that models can accurately label data, so in the interest of time and cost efficiency, we used an inter-model scheme to label the remaining notes (outlined in the Methods). The variability in the large dataset further underscores the necessity of our intelligent routing system to optimize performance by selecting the best model for each specific SDOH classification task (Figure \ref{figure:total_plot}).

\begin{figure}[ht]
\centering
\includegraphics[width= \linewidth]{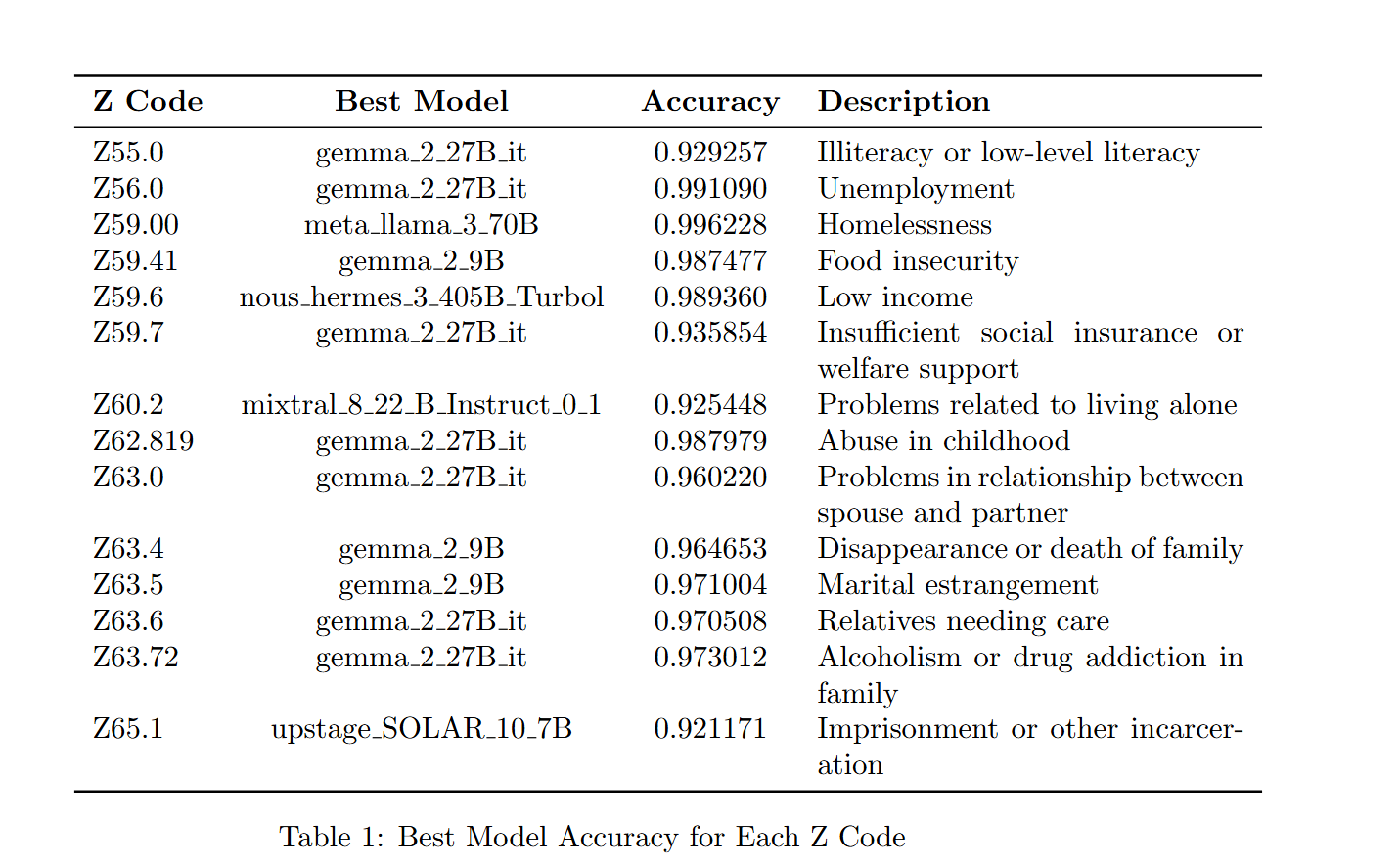}
\caption{\textbf{Router Chooses Best Model Each Code}. Shown above are the router chosen models for each SDOH code, along with its accuracy.}
\label{figure:total_model_chosen}
\end{figure}

As shown in Figure \ref{figure:total_model_chosen}, open-source models exhibit significant differential performance across medical notes. Some models, such as meta-llama-3-70B and gemma-2-27B-it, performed consistently well across multiple codes, while others showed specialized strengths for specific classifications, such as upstage-SOLAR-10.7B for imprisonment or other incarceration (Z65.1).

\begin{figure}[ht]
\centering
\includegraphics[width= \linewidth]{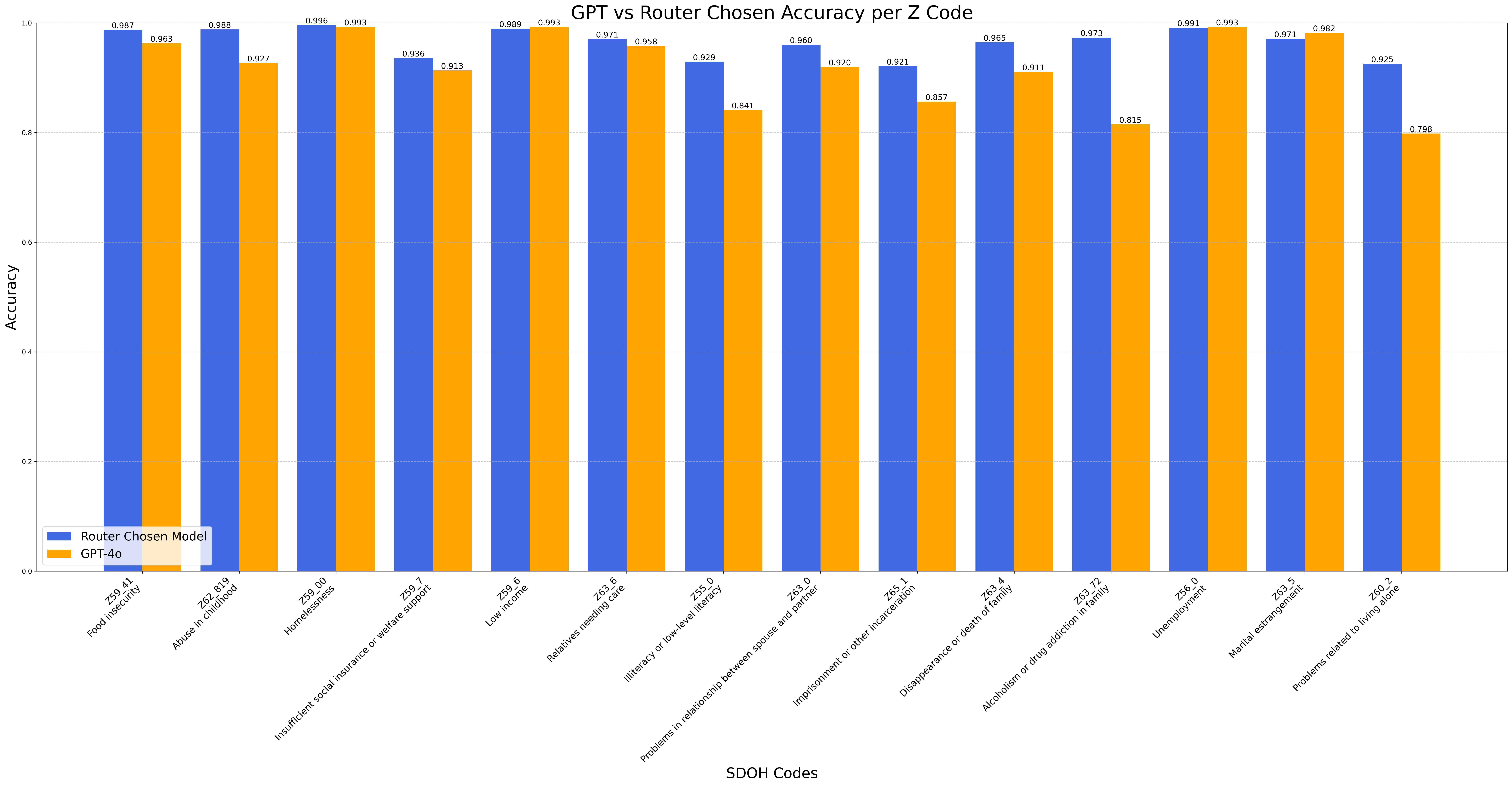}
\caption{\textbf{Router Chosen Model Outperforms GPT on SDOH coding task}. Expanding our dataset to 8000 medical notes and 13 SDOH codes, we see that in almost all the codes, our routing system of open-source models either performs comparably, or outperforms GPT-4o}
\label{figure:final_comparison}
\end{figure}

After implementing our intelligent routing system, we observed a significant improvement in accuracy compared to a single model approach. The router model achieved an average accuracy of 96.4\% across all SDOH codes, compared to 91.9\% for GPT-4o (Figure \ref{figure:final_comparison}. Additionally, the router demonstrated superior performance in complex classification tasks where GPT-4o struggled. For instance, the router outperformed classification accuracy by over 10\% for SDOH factors such as alcoholism or drug addiction in the family (Z63.72) and problems related to living alone (Z60.2), highlighting its ability to leverage specialized models for specific tasks.

These findings emphasize the value of an ensemble-based routing strategy in clinical natural language processing (NLP) applications. By dynamically selecting the most effective model for each classification task, our router system optimally balances performance across a range of SDOH codes, offering a scalable, cost-effective, and privacy-conscious alternative to closed-source models. This approach demonstrates the feasibility of high-performance medical text classification using exclusively open-source language models, paving the way for future expansion into additional clinical NLP applications.

\section{Conclusion}

To determine if we could apply large language models (LLMs) to identify relevant social determinants of health (SDOH)-related Z-codes in the unstructured free text of medical notes from a publicly-available, deidentified data set, we developed a novel architecture utilizing an intelligent routing of inputs to optimal language models that is further augmented by the generation of synthetic medical records data. Here, we demonstrate the high performance of our intelligent router, which utilizes the power of multiple open-source accuracy to perform slightly better than state-of-the-art models such as GPT-4o. We also introduced a novel synthetic data generation and validation scheme, which can help overcome LLM testing and training challenges related to data paucity, data privacy, and data quality issues.  The high accuracy we achieved was without any prompt-tuning or fine-tuning that could potentially boost performance further. We also plan to expand this system to account for more SDOH codes. This novel approach holds promise for health systems who may need ways to convert unstructured free text from a medical note into actionable, SDOH data that can systematically identify patients with health-related social needs who require intervention.

\textbf{Keywords:} Large language models, artificial intelligence, social determinants of health, electronic health
records, Z codes, intelligent routing.

\clearpage

\bibliographystyle{unsrt}
\bibliography{pre_print}

@techreport{solar2010conceptual,
  author      = {{Orielle Solar and Alec Irwin}},
  title       = {A conceptual framework for action on the social determinants of health},
  institution = {World Health Organization},
  address     = {Geneva, Switzerland},
  year        = {2010},
  note        = {Social determinants of health discussion paper 2 (policy and practice)}
}

@article{waltman2024margins,
  author  = {{Belinda Waltman}},
  title   = {The margins matter},
  journal = {JAMA},
  volume  = {331},
  number  = {1},
  pages   = {23},
  month   = jan,
  year    = {2024}
}

@techreport{cms2023zcode,
  author      = {{Centers for Medicare \& Medicaid Services}},
  title       = {CMS office of minority health z-code resource},
  institution = {Centers for Medicare \& Medicaid Services},
  year        = {2023},
  type        = {Technical report}
}

@article{clusmann_future_2023,
  author  = {{Clusmann, J., Kolbinger, F.R., Muti, H.S. \emph{et~al.}}},
  title   = {The future landscape of large language models in medicine},
  journal = {Commun Med},
  volume  = {3},
  pages   = {141},
  year    = {2023},
  note    = {\url{https://doi.org/10.1038/s43856-023-00370-1}}
}

@article{omiye_large_2024,
  author  = {{Omiye JA, Gui H, Rezaei SJ, Zou J, Daneshjou R}},
  title   = {Large Language Models in Medicine: The Potentials and Pitfalls : A Narrative Review},
  journal = {Ann Intern Med.},
  volume  = {177},
  number  = {2},
  pages   = {210--220},
  month   = feb,
  year    = {2024},
  note    = {doi: 10.7326/M23-2772}
}

@article{mehandru_evaluating_2024,
  author  = {{Mehandru, N., Miao, B.Y., Almaraz, E.R. \emph{et~al.}}},
  title   = {Evaluating large language models as agents in the clinic},
  journal = {npj Digit. Med.},
  volume  = {7},
  pages   = {84},
  year    = {2024},
  note    = {\url{https://doi.org/10.1038/s41746-024-01083-y}}
}

@article{guevara_large_2024,
  author  = {{Guevara, M., Chen, S., Thomas, S. \emph{et~al.}}},
  title   = {Large language models to identify social determinants of health in electronic health records},
  journal = {npj Digit. Med.},
  volume  = {7},
  pages   = {6},
  year    = {2024},
  note    = {\url{https://doi.org/10.1038/s41746-023-00970-0}}
}

@article{lybarger_annotating_2021,
  author  = {{Lybarger K, Ostendorf M, Yetisgen M}},
  title   = {Annotating social determinants of health using active learning, and characterizing determinants using neural event extraction},
  journal = {J Biomed Inform.},
  volume  = {113},
  pages   = {103631},
  month   = jan,
  year    = {2021},
  note    = {doi: 10.1016/j.jbi.2020.103631}
}

@misc{touvron_llama_2023,
  author = {{Hugo Touvron, Thibaut Lavril, Gautier Izacard, \emph{et~al.}}},
  title  = {LLaMA: Open and Efficient Foundation Language Models},
  year   = {2023},
  note   = {Version Number: 1. arXiv:2302.13971v1 [cs.CL].}
}

@misc{jiang_mistral_2023,
  author = {{Albert Q. Jiang, Alexandre Sablayrolles, Arthur Mensch, \emph{et~al.}}},
  title  = {Mistral 7B},
  year   = {2023},
  note   = {Version Number: 1. arXiv:2310.06825 [cs.CL].}
}

@misc{hari_herd_2023,
  author = {{Surya Narayanan Hari and Matt Thomson}},
  title  = {Herd: Using multiple, smaller LLMs to match the performances of proprietary, large LLMs via an intelligent composer},
  year   = {2024},
  note   = {Version Number: 2. arXiv:2310.19902v2 [cs.AI]}
}

@misc{hari_tryage_2023,
  author = {{Surya Narayanan Hari and Matt Thomson}},
  title  = {Tryage: Real-time, intelligent Routing of User Prompts to Large Language Models},
  year   = {2023},
  note   = {Version Number: 2. arXiv:2308.11601v2 [cs.LG].}
}

@article{cook_quality_2021,
  author  = {{Cook LA, Sachs J, Weiskopf NG}},
  title   = {The quality of social determinants data in the electronic health record: a systematic review},
  journal = {J Am Med Inform Assoc.},
  volume  = {29},
  number  = {1},
  pages   = {187--196},
  year    = {2021},
  note    = {doi: 10.1093/jamia/ocab199}
}

@misc{wang_openchat_2024,
  author = {{Guan Wang, Sijie Cheng, Xianyuan Zhan, Xiangang Li, Sen Song, and Yang Liu}},
  title  = {Open-Chat: Advancing Open-source Language Models with Mixed-Quality Data},
  month  = mar,
  year   = {2024},
  note   = {arXiv:2309.11235 [cs].}
}

@misc{bommasani_opportunities_2021,
  author = {{Rishi Bommasani, Drew A. Hudson, Ehsan Adeli, \emph{et~al.}}},
  title  = {On the Opportunities and Risks of Foundation Models},
  year   = {2021},
  note   = {Version Number: 3. arXiv:2108.07258v3 [cs.LG].}
}

@misc{jin_dataless_2022,
  author = {{Xisen Jin, Xiang Ren, Daniel Preotiuc-Pietro, and Pengxiang Cheng}},
  title  = {Dataless Knowledge Fusion by Merging Weights of Language Models},
  year   = {2025},
  note   = {Version Number: 6. arXiv:2212.09849v6 [cs.CL].}
}

@misc{johnson_mimic-iii_2015,
  author = {{Alistair Johnson, Tom Pollard, and Roger Mark}},
  title  = {MIMIC-III Clinical Database},
  year   = {2015}
}

\end{document}